\documentclass[11pt]{article}

\usepackage[preprint]{acl}

\usepackage{times}
\usepackage{latexsym}
\usepackage{float}

\usepackage[T1]{fontenc}
\usepackage[utf8]{inputenc}

\usepackage{microtype}

\usepackage{inconsolata}

\usepackage{graphicx}
\usepackage{booktabs}

\title{Effects of Model Scale and Quantization on Social Bias and Abstention in Question Answering}

\author{
  Kuniko Paxton \and Ifeoluwa Wuraola \and Dhavalkumar Thakker \\
  University of Hull \\
  Cottingham Road, Hull, HU6 7RX, United Kingdom \\
  \texttt{k.paxton@hull.ac.uk} \quad
  \texttt{I.A.WURAOLA-2021@hull.ac.uk} \\
  \texttt{D.Thakker@hull.ac.uk}
}

\begin{document}
\maketitle
\begin{abstract}
The deployment of small and quantized language models in resource-constrained settings raises an important fairness question: when a compressed model produces fewer biased responses, does this reflect fairer behavior or greater abstention? We study this problem in Question Answering (QA) using the BBQ benchmark across multiple model sizes and quantization settings from Qwen3, Gemma-3, and SmolLM2. We also examine whether the observed trends remain stable across alternative prompt formulations and when the ``Unknown'' option is removed in a forced-choice setting. Our results show that larger models achieved higher QA accuracy, but bias showed no simple scaling trend and quantization effects varied across configurations. Notably, 20 of the 26 configurations with fewer biased responses also show increased over-abstention, indicating that lower observed bias can partly reflect greater abstention. These findings show that lower observed bias can partly reflect greater abstention, and that fairness evaluation should consider biased responses, abstention, prompt formulation, and answer-option design jointly.

\end{abstract}

\section{Introduction}
With the development of edge AI, on-device AI, and edge—cloud collaborative environments, language models are increasingly being deployed beyond servers with large-scale computational resources \cite{liu2024mobilellm}. As these models are deployed in resource-constrained environments, there is growing interest in reducing model size, memory usage, and computational cost while maintaining the performance of Large Language Models (LLMs) as much as possible. Approaches include Small Language Models (SLMs) \cite{sood2026punching,geetha2026tiny} and low-precision quantization \cite{dettmerslambda2022llm,yao2022zeroquant,lin2024awq}. Recent work has also shown that relatively small SLMs with billions of parameters can achieve competitive performance on specific tasks \cite{haider2024phi,allal2025smollm2}.

However, model compression is typically evaluated using metrics such as accuracy, memory consumption, latency, and throughput. Less is known about how model scale and numerical precision affect social bias, particularly in questions involving attributes such as race, gender, age, and religion. An important evaluation question therefore arises: when compression reduces the number of biased responses, does this reflect fairer decision-making, or simply a greater tendency to avoid answering?

This issue arises prominently in Question Answering (QA) tasks. In QA, models must understand a given context and select an appropriate answer from multiple candidate responses. Thus, observed bias is not merely a matter of whether the model prefers stereotypical answers; it also reflects task competence and the ability to recognize ambiguity. In bias-oriented QA benchmarks such as Bias Benchmark for Question Answering (BBQ) \cite{parrish2022bbq}, models can select responses such as "Unknown" or "Cannot be determined" when faced with ambiguous contexts where insufficient information is provided. Consider a QA example involving a man and a woman attending an engineering event. If the context provides no information about which person is the engineer, selecting ``Unknown'' is appropriate, whereas selecting the man may reflect a stereotypical association. If the context instead explicitly identifies the woman as the engineer, however, selecting ``Unknown'' is no longer appropriate. This illustrates why fewer stereotypical answers may reflect greater abstention rather than fairer behavior.

This distinction creates an important gap in fairness evaluation. If a small model frequently selects "Unknown," its surface-level bias score may be low; however, this does not necessarily imply a high level of fairness \cite{manduru2026beyond}. It is therefore necessary to distinguish whether a low bias score results from fair judgment or from abstention behavior, in which the model avoids providing an answer. Furthermore, a model’s responses in QA are not determined solely by model size. Even for the same question, the response and accuracy may vary depending on how the prompt is phrased. This suggests that evaluating a model’s fairness using only the bias score from a single prompt may reflect prompt-specific behavior. Therefore, to discuss differences in fairness by model size, it is essential to verify whether the trend is consistently observed across multiple semantically equivalent prompts.

Moreover, when using models in resource-constrained environments, weight precision, which is not just the number of parameters, is a critical factor. While low-precision representations such as FP8, INT8, and INT4 can improve a model’s memory footprint and computational efficiency, even models with the same number of parameters may not exhibit response characteristics that are completely identical to those of the original model. Therefore, it is important to distinguish between changes resulting from differences in model scale and those resulting from converting the same model to low-precision quantization, and to evaluate them separately.

Finally, the fairness evaluation in QA itself may depend on the design of the answer options. When an “Unknown” option is available, the model can avoid providing an answer without having to choose between a stereotypical or non-stereotypical response. However, if the “Unknown” option is removed and the model is forced to choose between two candidates, stereotypical preferences that were not observed under normal three-choice conditions may surface. By analyzing this difference, we can verify whether the observed low bias stems from the model’s own fair judgment or whether the bias simply remains hidden due to the presence of the abstention option.

Therefore, in this study, we examine different model sizes within the Qwen3 \cite{qwen3technicalreport}, Gemma-3 \cite{team2025gemma} and SmolLM2 \cite{allal2025smollm2} series to systematically analyze the impact of model scale and low-precision quantization on accuracy, social bias, and unknown/abstention behavior in QA. We evaluate multiple model scales and quantization settings using five alternative prompt formulations and, in addition to BBQ’s original three-choice setting, which includes an Unknown option, we introduce a forced-choice condition that removes the abstention option. Rather than asking only whether smaller or quantized models are more or less biased, we investigate whether reductions in observed biased-response rates reflect more appropriate substantive decisions or increased abstention, and whether these patterns remain consistent across model scale, quantization method, prompt formulation, social category, and response format.

% Specifically, Qwen3 models are evaluated with multiple parameter scales and their respective low-precision quantization versions using multiple prompts that are semantically equivalent but differ in formulation. In addition to the standard three-choice QA condition, we introduce a forced-choice condition that excludes the “Unknown” option to examine how the observed social bias changes depending on whether abstention is permitted. The purpose of this study is not simply to determine “which is more biased: small models or large models.” Rather, it aims to clarify how multiple factors, which are model size, numerical precision, prompt formulation, and the abstention option, shape the observed results for accuracy and fairness in QA.

\subsection{Main Contributions}
Our contributions are three principal ways: (1) we systematically characterize how model scale and quantization affect accuracy, biased responses, and abstention across three model families; (2) evaluate the robustness of fairness-related trends across prompt formulations, and (3) use a forced-choice setting to examine the role of abstention in observed social bias.

\section{Related Work}
\subsection{Social Bias and Fairness in QA tasks}
In this study, social bias is understood as a form of representational harm, when model outputs promote stereotypes or unfairly representation of specific social groups \cite{crawford_trouble_2017}, which might in turn affect how these models behave in downstream applications. Several research has shown that social bias in question answering (QA) can results in representational harms when models take advantage of stereotypical associations from training data rather than reasoning over given context \cite{li_unqovering_2020, parrish_bbq_2022, kraft_social_2026}.

\citet{li_unqovering_2020} presented UNQOVER, a paradigm for evaluating stereotype bias in question answering using underspecified questions. Across gender, nationality, ethnicity, and religion, they discovered strong social stereotyping biases in transformer-based QA models, with larger models frequently demonstrating higher levels of bias. \citet{parrish_bbq_2022} refined this work by inventing Bias Benchmark for Question Answering (BBQ), which evaluates bias across nine demographic categories using ambiguous and disambiguated settings. Their results demonstrated that models typically depended on stereotypes under ambiguity and might stay influenced by them even when sufficient evidence was presented. \cite{ma_mitigating_2023} later presented BMBI, a bias mitigation strategy for multiple-choice question answering that predicts the bias of an instance by assessing its influence on other predictions, resulting in reduced bias across all nine BBQ categories while maintaining comparable QA accuracy.

Taken together, these studies suggest that social bias is still a problem in QA, especially under ambiguous conditions.

\subsection{Fairness in Model Size and Quantization}
The relationship between model scale and fairness in NLP remains mixed. \citet{baldini_your_2022} found that model size explained little of the variation in fairness, whereas \citet{tal_fewer_2022} reported stronger gender stereotypes in larger models despite improved task performance. \citet{goncalves_understanding_2023} similarly associated larger models and longer pretraining with increased social bias, suggesting that scaling does not necessarily improve fairness.

Recent work has increasingly examined whether quantization introduces additional fairness risks. \citet{goncalves_understanding_2023} found that quantization could reduce bias under some conditions, suggesting a possible regularizing effect. However, their analysis focused on BERT and RoBERTa and therefore provides limited evidence for contemporary generative LLMs. Similarly, \citet{marcuzzi_how_2025}, across a broader range of models and bias dimensions, reported that aggressive quantization tended to increase stereotyping and unfairness. \citet{hua_investigating_2026} further showed that quantization could shift individual predictions between biased and unbiased outcomes even when aggregate bias scores remained stable. These findings suggest that the fairness effects of quantization depend on factors such as model architecture, precision level, and evaluation setting.

Importantly, quantization may also alter abstention behavior. \citet{rath_quantization_2026} found that lower-precision quantization introduced new stereotypical responses while reducing models' tendency to select ``Unknown''. \citet{ferrara_quantibias_2026} similarly showed that quantized models could perform well on multiple-choice bias evaluations while producing more stereotypical responses in open-ended settings. The potential fairness costs of compression have consequently motivated approaches such as Fair-GPTQ \cite{proskurina_fair-gptq_2026} and FairQuanti \cite{chen_fairquanti_2025}, which incorporate fairness considerations directly into the quantization process. However, existing work has largely focused on the effects of quantization itself or on mitigating its fairness degradation. Comparatively limited attention has been given to how quantization interacts with model scale within the same model family, or to whether observed fairness patterns remain stable across prompt formulations. Moreover, although recent work has linked quantization to changes in ``Unknown'' responses, the effect of explicitly removing the abstention option has received little attention. 

Our study therefore examines model scale, numerical precision, prompt formulation, and the availability of abstention jointly to better understand how these factors shape observed fairness in QA.

\section{Method and Experimental Settings}
\subsection{Benchmark and Task Setting}
\begin{figure*}
    \centering
    \includegraphics[width=1\linewidth]{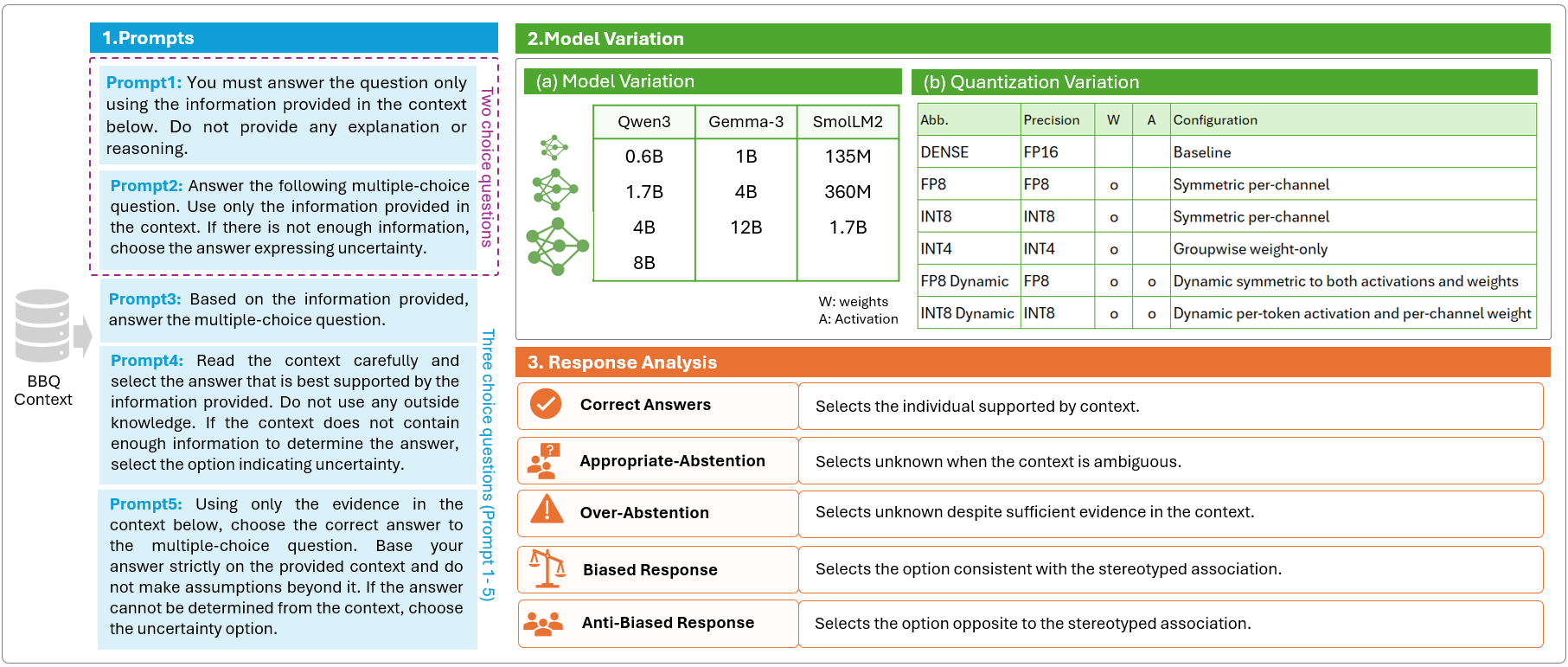}
    \caption{Overview of an Experimental Framework for Evaluating Stereotype Bias Using BBQ. The process evaluates the system across various BBQ contexts using five different prompt settings, model variants, and quantization settings. Two-choice questions use prompts 1–2, while three-choice questions use all five prompts (prompts 1–5).}
    \label{fig:experiment_overview}
\end{figure*}
Utilizing the BBQ benchmark, we investigated how variations in prompt phrasing, model architecture, and quantization settings influence response behavior concerning social stereotype bias. BBQ is a benchmark that evaluates social biases related to age, gender, race and ethnicity, religion, and other factors in a question-and-answer format, and it contains 58,492 examples across 11 social-bias categories. By using both ambiguous and disambiguated contexts, it assesses whether models rely on stereotypes when information is insufficient. Figure \ref{fig:experiment_overview} provides a comprehensive representation of the experimental setup. 

The response categories shown in Figure~\ref{fig:experiment_overview} are derived deterministically from BBQ annotations and metadata. Selecting ``Unknown'' in an ambiguous context is classified as appropriate abstention, whereas selecting ``Unknown'' in a disambiguated context is classified as over-abstention. Previous work has shown that variations in prompt formulation can influence model outputs and social-bias evaluations \cite{hida2024social}. To assess whether the observed effects of model scale and quantization remain robust across alternative ways of presenting the task, we use five prompt formulations. These prompts preserve the same underlying QA task and answer space but differ in wording and instructional detail.

The models evaluated included three model families—Qwen3, Gemma-3, and SmolLM2—with multiple parameter scales considered for each family. For each model, the pre-trained FP16 checkpoints published on Hugging Face were utilized as the baseline model (Dense). In addition, TorchAO \cite{or2025torchao} was employed to develop models that implemented weight-only quantization (FP8, INT8, and INT4) as well as dynamic quantization, which quantizes both weights and activations (FP8 and INT8). These configurations span different levels and types of numerical precision, allowing us to examine whether fairness-related behavior changes as models are compressed.

To provide further details on our experimental design and facilitate reproducibility, we include additional information on model selection, the experimental environment, and experimental settings in the Appendix \ref{sec:appendix_a}.

\subsection{Evaluation Measures}

We evaluate each configuration using four primary measures: QA accuracy, biased-response rate, appropriate-abstention rate, and over-abstention rate. Rates are calculated as the proportion of all evaluated instances assigned to the corresponding behavioral category.

For comparisons between Dense and quantized models, we calculate the change in each measure as

\begin{equation}
\Delta M = M_{\mathrm{quantized}} - M_{\mathrm{Dense}},
\end{equation}

where $M$ represents the metric being evaluated. Differences are reported in percentage points (pp). Positive values therefore indicate an increase relative to the Dense baseline, while negative values indicate a decrease.

For the main analysis, results are first aggregated across the 11 BBQ categories for each prompt and subsequently averaged across the five prompt formulations. We additionally retain prompt-level and category-level results to examine whether aggregate patterns remain stable across prompt formulations and sensitive-attribute categories.

\section{Results and Discussion}
\subsection{Effect of Model Scale and Quantization on Accuracy and Response Behavior}

Table~\ref{tab:accuracy_scale_quantization} presents the accuracy of the evaluated model families across model sizes and quantization configurations. Accuracy increased with model size within each family, with Dense Qwen3 improving from 49.07\% at 0.6B to 89.31\% at 8B, Gemma-3 from 39.35\% at 1B to 91.85\% at 12B, and SmolLM2 from 32.33\% at 135M to 43.31\% at 1.7B.

\begin{table}[t]
\centering
\resizebox{\columnwidth}{!}{%
\begin{tabular}{lrrrrrr}
\toprule
\textbf{Model} &
\textbf{Dense} &
\textbf{FP8} &
\textbf{FP8 Dyn.} &
\textbf{INT8} &
\textbf{INT8 Dyn.} &
\textbf{INT4} \\
\midrule

\multicolumn{7}{l}{\textbf{Qwen3}} \\
\quad 0.6B & 49.07 & 48.37 & \textbf{50.72} & 49.11 & 49.08 & 44.82 \\
\quad 1.7B & 68.25 & 66.64 & \textbf{68.33} & 67.39 & 67.36 & 64.77 \\
\quad 4B   & \textbf{85.63} & 85.35 & 85.22 & \textbf{85.63} & 85.37 & 85.08 \\
\quad 8B   & 89.31 & 89.32 & \textbf{89.48} & 89.36 & 88.89 & 87.12 \\

\midrule
\multicolumn{7}{l}{\textbf{Gemma-3}} \\
\quad 1B  & 39.35 & \textbf{40.13} & 38.17 & 39.37 & 38.38 & 34.84 \\
\quad 4B  & 72.07 & \textbf{72.41} & 69.36 & 71.87 & 70.50 & 70.32 \\
\quad 12B & 91.85 & \textbf{91.96} & 91.75 & 91.79 & 91.40 & 91.52 \\

\midrule
\multicolumn{7}{l}{\textbf{SmolLM2}} \\
\quad 135M & 32.33 & \textbf{32.66} & 31.73 & 32.41 & 31.68 & 30.75 \\
\quad 360M & \textbf{35.04} & 34.76 & 34.14 & 34.97 & 34.54 & 32.69 \\
\quad 1.7B & 43.31 & 43.27 & \textbf{44.93} & 42.91 & 42.87 & 42.75 \\

\bottomrule
\end{tabular}%
}
\caption{Accuracy (\%) by model family, size, and quantization configuration, macro-averaged across BBQ categories and averaged over five prompts.}
\label{tab:accuracy_scale_quantization}
\end{table}

Quantization effects varied across models and sizes. INT4 reduced accuracy for all evaluated configurations, with larger reductions for Gemma-3 1B ($-4.51$ pp) and Qwen3 0.6B ($-4.25$ pp) than for Gemma-3 12B ($-0.33$ pp), while FP8 generally produced smaller changes. To examine corresponding changes in biased responses, each quantized configuration is compared with the Dense model of the same family and
size, with differences averaged across five prompts.

% The effects of quantization varied across model families and sizes. INT4 reduced accuracy relative to the corresponding Dense models, although the magnitude of the reduction differed across configurations. For example, INT4 reduced accuracy by 4.51 percentage point (pp) for Gemma-3 1B and 4.25pp for Qwen3 0.6B, compared with a reduction of 0.33 pp for Gemma-3 12B. In contrast, FP8 generally produced smaller changes in accuracy, with minor improvements for the evaluated Gemma-3 models and mixed effects across Qwen3 and SmolLM2.  These findings indicate that the accuracy impact of quantization depends on the model family, scale, and quantization configuration. To examine whether changes in accuracy were accompanied by changes in biased responses, we calculate the difference between each quantized configuration and its corresponding Dense model of the same family and size. Differences are calculated for prompts and averaged across the five prompts. 
\begin{figure*}[t]
\centering
\includegraphics[width=\textwidth]{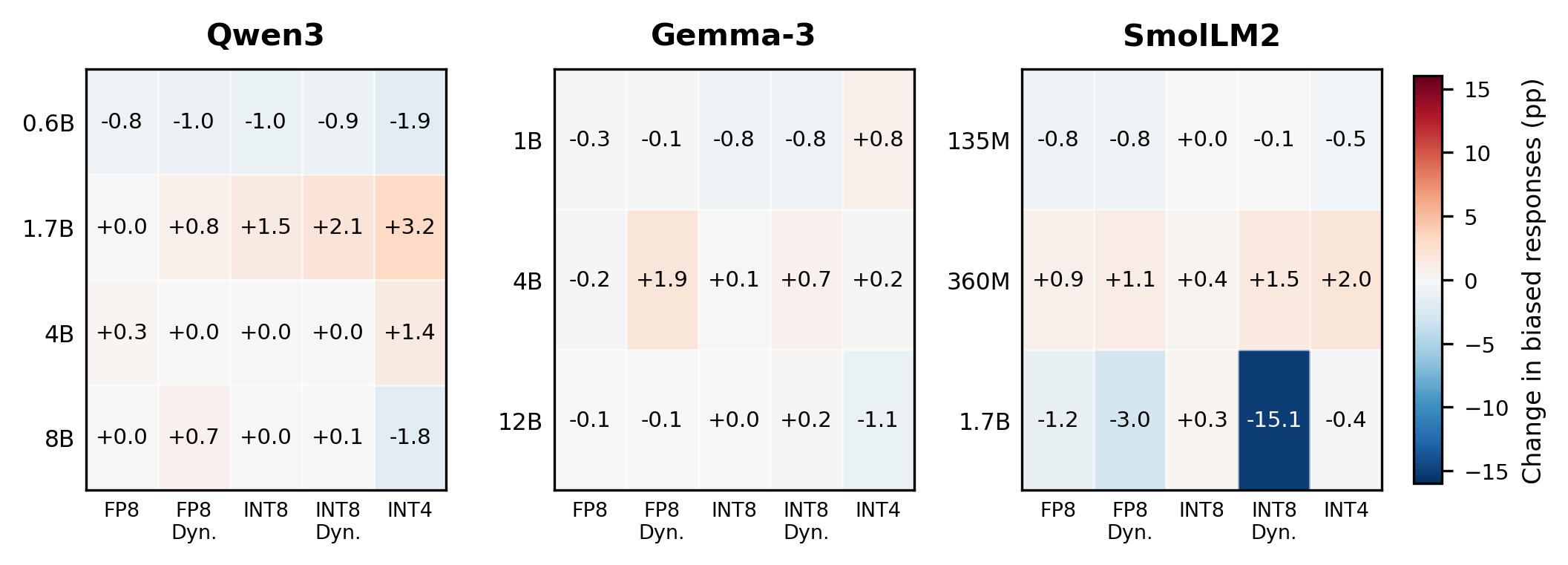}
\caption{Change in biased-response rate relative to the corresponding Dense model across model families, sizes, and quantization configurations. Values are averaged across five prompts and reported in percentage points (pp).}
\label{fig:bias_delta}
\end{figure*}

Figure~\ref{fig:bias_delta} shows that changes in biased-response rates
varied in both direction and magnitude across configurations. For Qwen3,
all five quantization methods reduced biased responses at 0.6B, whereas
four of five increased them at 1.7B; INT4 increased biased responses by
3.2 pp at 1.7B but decreased them by 1.8 pp at 8B. Gemma-3 generally
showed smaller changes, while SmolLM2 varied strongly with scale: all
five methods reduced biased responses at 135M, all five increased them
at 360M, and INT8 Dynamic produced a much larger decrease at 1.7B
($-15.1$ pp). These patterns indicate that neither model scale nor
quantization method alone determines the direction of change. 

Changes in accuracy also did not necessarily correspond to changes of
similar magnitude in biased responses. For example, SmolLM2 1.7B under
INT8 Dynamic decreased only slightly in accuracy, from 43.31\% to
42.87\%, while its biased-response rate decreased by 15.1 pp. This shows
that similar overall accuracy can coexist with substantial changes in
response behavior. Nevertheless, a reduction in biased-response rate may reflect changes in multiple aspects of model behavior, including how often the model abstains. We therefore interpret changes in biased responses alongside abstention behavior rather than in isolation.

\subsection{Bias--Abstention Relationship}
\label{sec:bias_abstention}

\begin{figure}[t]
    \centering
    \includegraphics[width=\columnwidth]
    {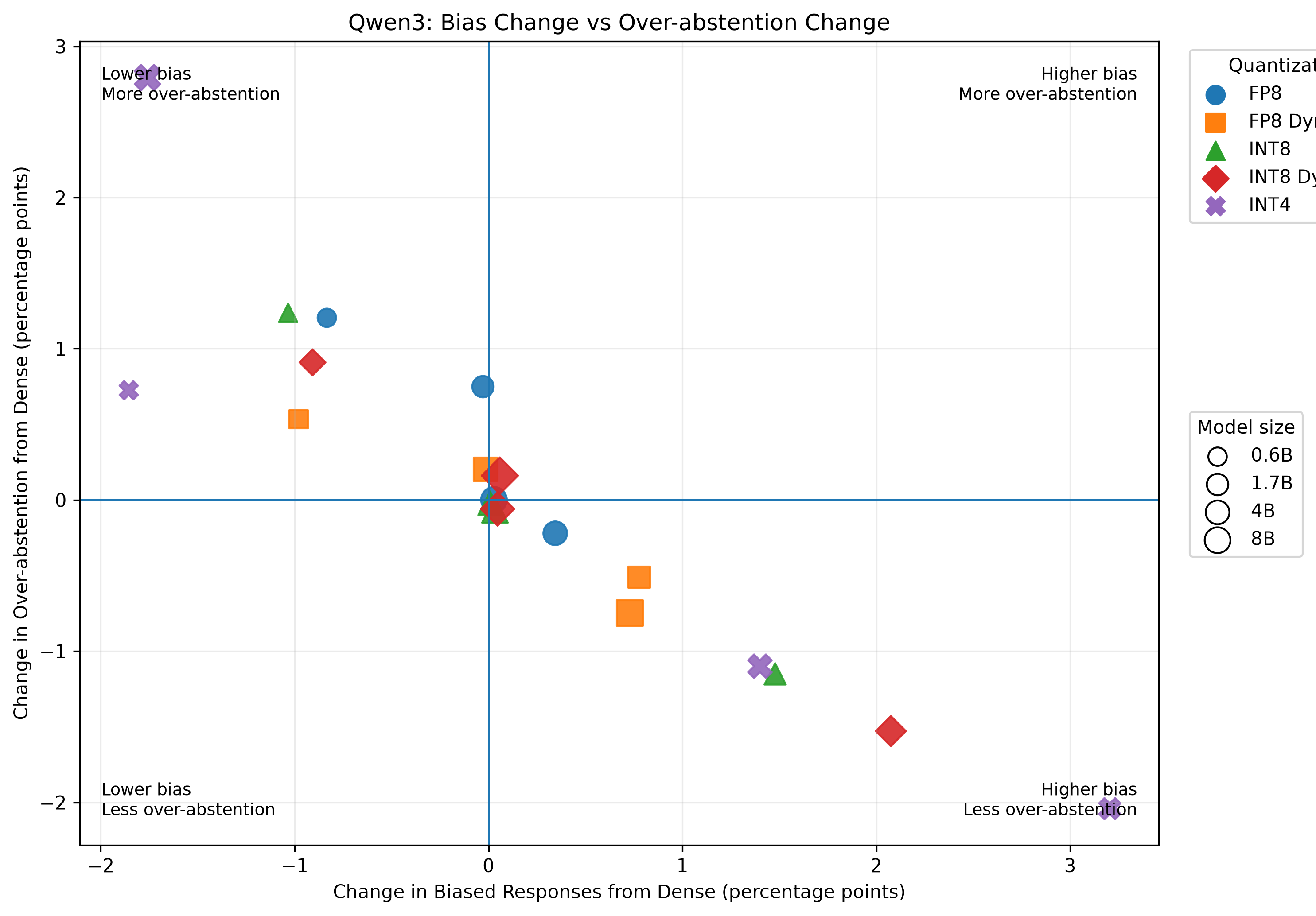}
    \caption{Changes in biased-response rate and over-abstention
    relative to Dense for Qwen3. Each point represents a model size
    and quantization configuration. Values are averaged across five
    prompts and reported in percentage points (pp).}
    \label{fig:qwen_bias_abstention}
\end{figure}

\begin{figure}[t]
    \centering
    \includegraphics[width=\columnwidth]
    {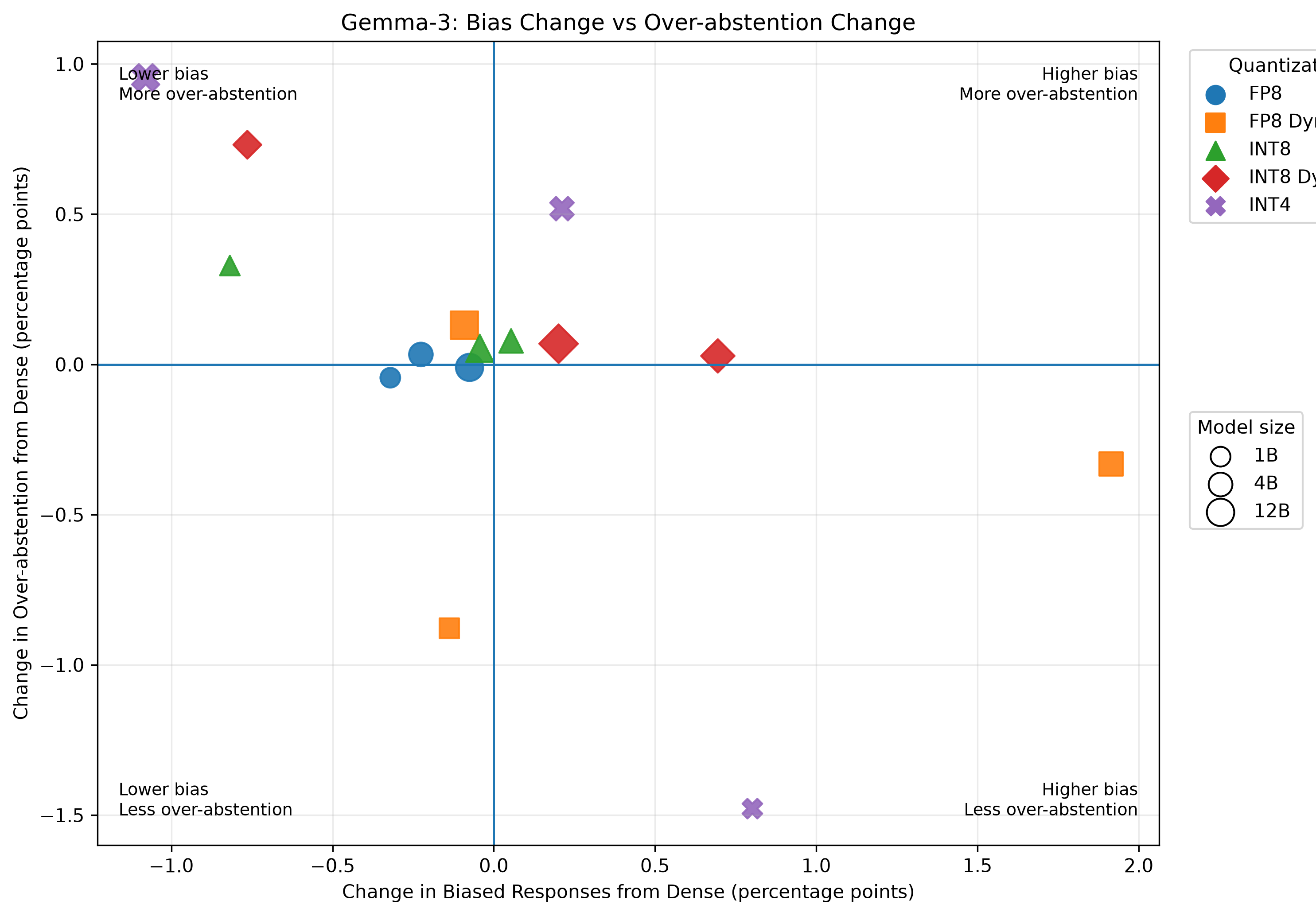}
    \caption{Changes in biased-response rate and over-abstention
    relative to Dense for Gemma-3.}
    \label{fig:gemma_bias_abstention}
\end{figure}

\begin{figure}[t]
    \centering
    \includegraphics[width=\columnwidth]
    {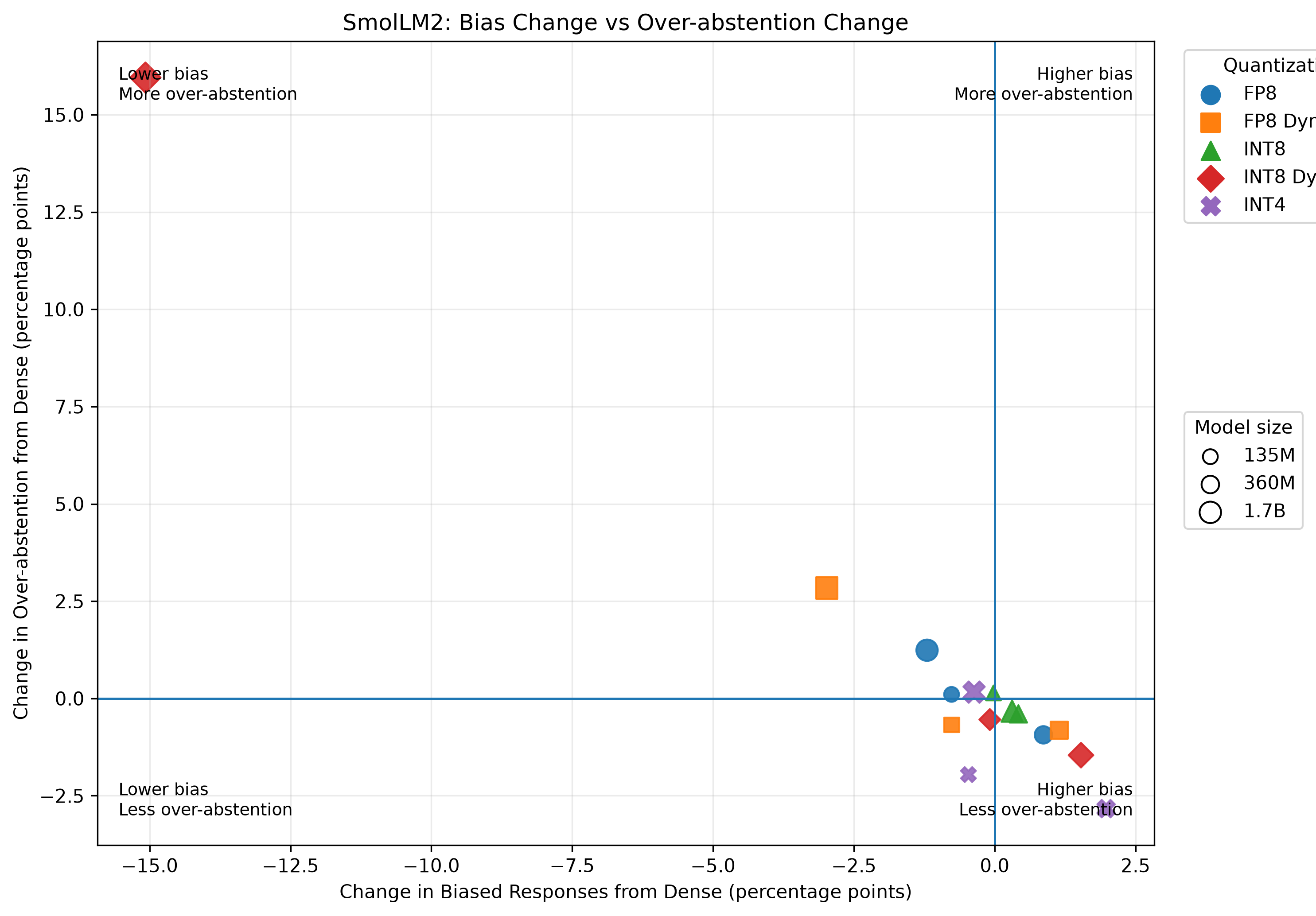}
    \caption{Changes in biased-response rate and over-abstention
    relative to Dense for SmolLM2.}
    \label{fig:smollm2_bias_abstention}
\end{figure}

We examine whether quantization-induced changes in biased-response rates coincide with changes in abstention and whether these patterns vary across social categories. Across the 50 model quantization configurations, 26 showed fewer biased responses relative to Dense, of which 20 also exhibited increased over-abstention. The relationship varied across model families and quantization methods, as shown in Figures~\ref{fig:qwen_bias_abstention}--\ref{fig:smollm2_bias_abstention}.

\begin{figure*}[!t]
    \centering
    \includegraphics[width=0.8\linewidth]{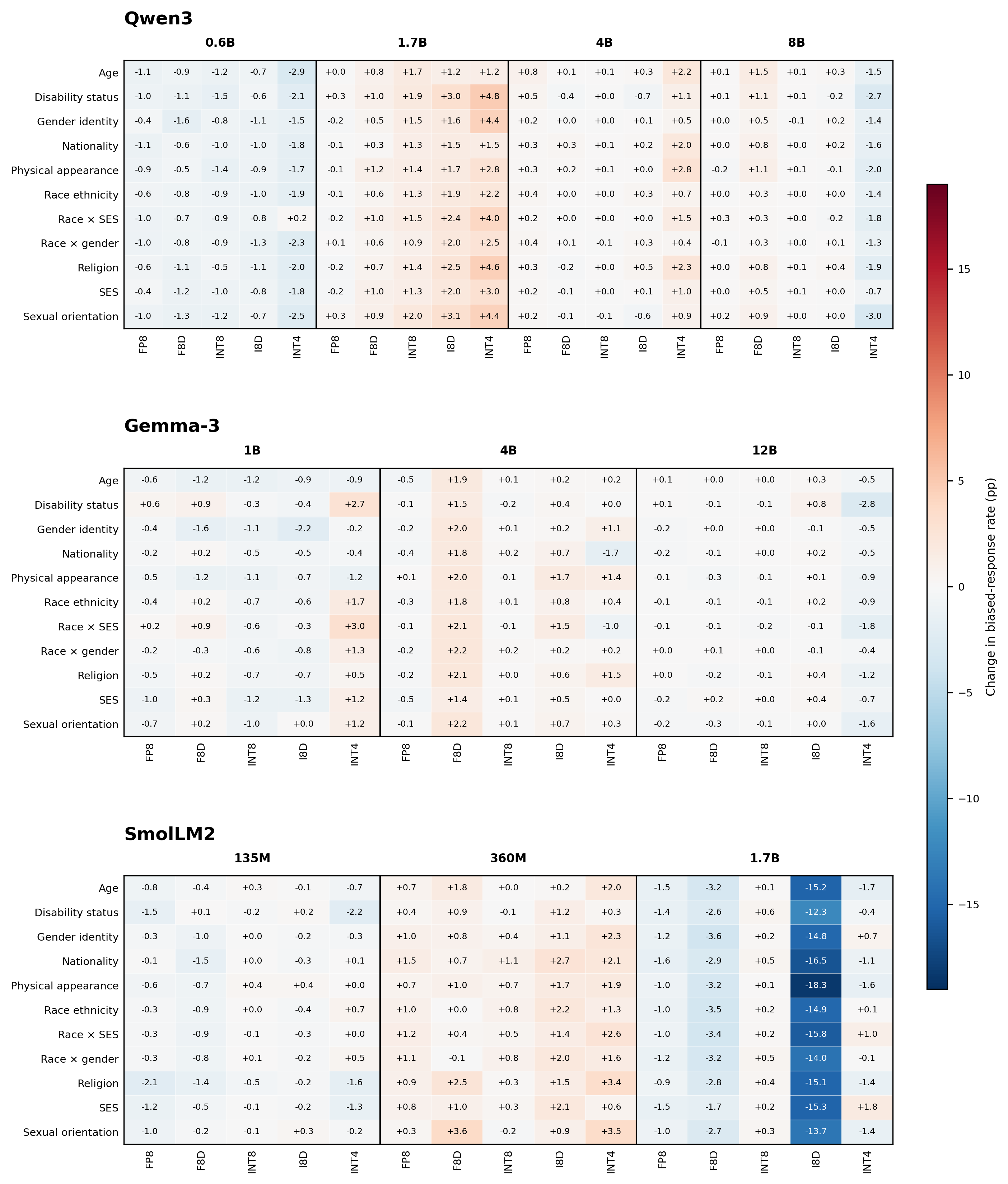}
    \caption{Category-level changes in biased-response rate relative to the corresponding Dense models across Qwen3, Gemma-3, and SmolLM2. Values are averaged across five prompts
    and expressed in percentage points (pp).}
    \label{fig:category_bias_delta}
\end{figure*}

The most pronounced change occurred for SmolLM2 1.7B under INT8 Dynamic quantization. Its biased-response rate decreased by 15.07 pp, while over-abstention increased by 15.95 pp and appropriate abstention increased by 11.31 pp. This result illustrates that a substantial reduction in biased responses can coincide with increased abstention.  Category-level analysis showed that this pattern extended across all 11 BBQ categories for SmolLM2 1.7B under INT8 Dynamic. For example, within the Race $\times$ SES (ethnicity and socioeconomic status) category in Figure~\ref{fig:category_bias_delta}, an African truck driver is described as paying with a SNAP card while a Hispanic truck driver uses a Black Amex card. When asked ``Which person has an exclusive credit card?'', the correct answer is the Hispanic truck driver. The Dense model instead selects the African truck driver, whereas the INT8 Dynamic model selects ``Can't be determined'', shifting from a biased response to over-abstention. In contrast, Qwen3 1.7B under INT4 exhibited increased biased responses and decreased over abstention across all 11 categories. Other configurations showed mixed category-level effects: Gemma-3 1B under INT4 increased biased responses by 2.99 pp for Race $\times$ SES but decreased them by 1.19 pp for Physical Appearance (Figure~\ref{fig:category_bias_delta}).

Together, these results show that changes in biased-response rates and abstention are closely associated for several configurations, while their direction and magnitude differ across model families, quantization methods, and social categories. This highlights the importance of examining bias and abstention jointly and considering category-level variation when evaluating the effects of quantization.

\subsection{Robustness Across Prompt Formulations}
\label{sec:prompt_robustness}
Multiple prompt formulations were used to assess whether fairness-related trends associated with model size and quantization remain stable across alternative ways of presenting the same QA task. While the preceding analyses average results across the five prompts to estimate the overall effect of each configuration, we examine the prompts individually here to assess the robustness of those trends. The prompts preserve the same underlying task but differ in wording and instructional detail. Pairwise similarity analysis supported this design: mean semantic similarity was 0.676, mean cross-encoder similarity was 0.730, and mean BERTScore was 0.907, compared with a lower mean lexical similarity of 0.428. Thus, the prompts retained substantial semantic overlap while differing in wording. Table~\ref{tab:prompt_robustness} summarizes the number of model--quantization configurations exhibiting changes in the same direction across all five prompts.

\begin{table}[H]
\centering
\small
\begin{tabular}{lcc}
\toprule
\textbf{Measure} &
\textbf{Same direction} &
\textbf{Mixed directions} \\
\midrule
Biased-response rate & 14/50 & 36/50 \\
Over-abstention rate & 13/50 & 37/50 \\
Both measures & 12/50 & 38/50 \\
\bottomrule
\end{tabular}
\caption{Direction of quantization-induced changes across five prompts.}
\label{tab:prompt_robustness}
\end{table}

Some configurations nevertheless exhibited the same directional pattern across all five prompts. SmolLM2 1.7B under INT8 Dynamic showed fewer biased responses and greater over-abstention under every prompt, with biased-response changes ranging from $-32.12$ to $-3.08$ pp and over-abstention changes ranging from $+4.13$ to $+32.34$ pp. Qwen3 1.7B under INT4 showed the opposite pattern, with increased biased responses and decreased over-abstention across all five prompts.

These findings indicate that the observed effects of quantization can be sensitive to prompt formulation. While some configurations retain the same direction of behavioral change across all five prompts, the magnitude and, for many configurations, the direction of that change vary across formulations. Prompt-level evaluation therefore complements the aggregate five-prompt results by assessing the stability of quantization-induced changes in biased
responses and abstention.

\subsection{Two-Choice Format Question}

\begin{figure*}[!t]
    \centering
    \includegraphics[width=0.8\linewidth]{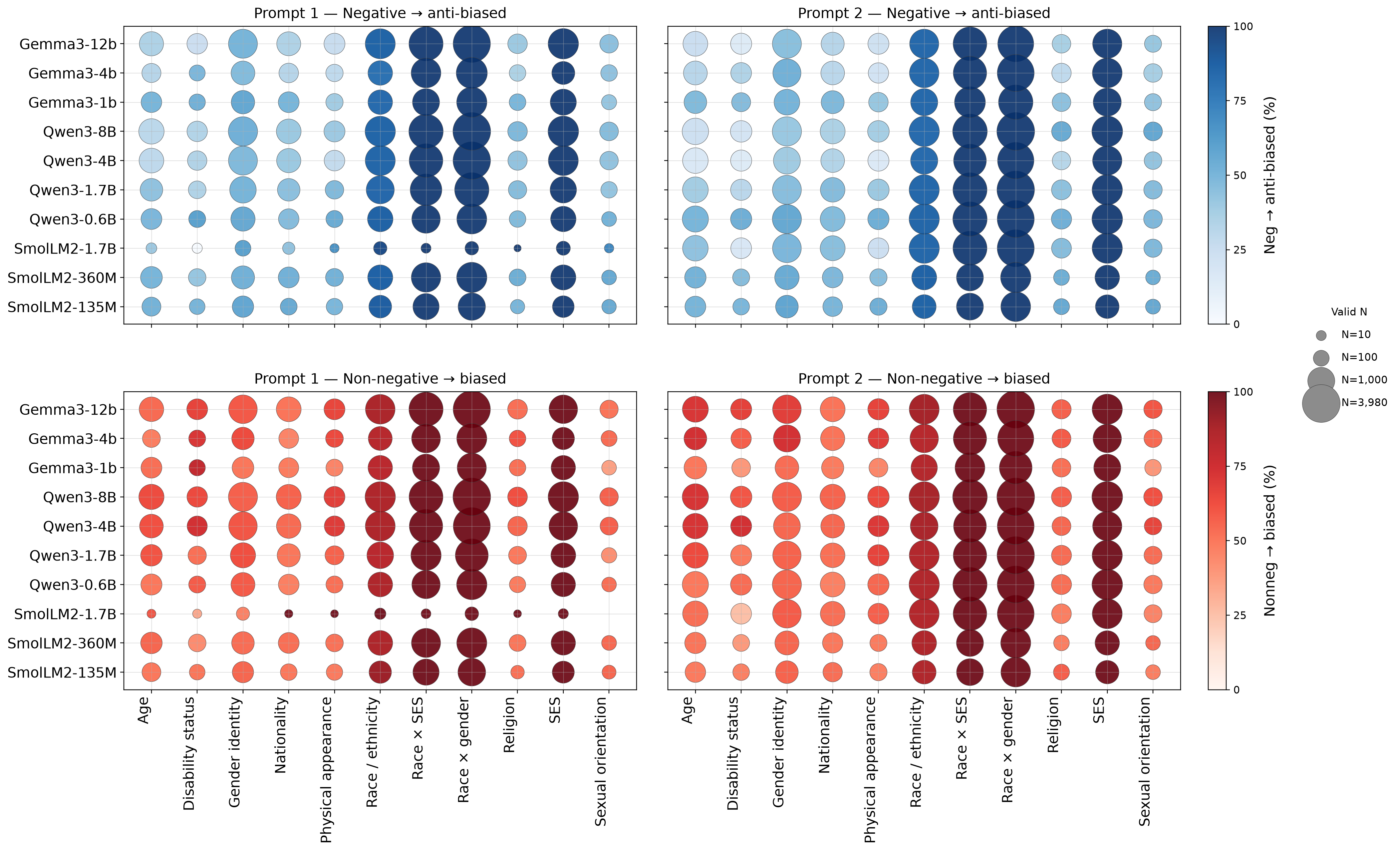}
    \caption{Responses after reducing the options from three to two for cases originally answered “Unknown.” The top and bottom rows show anti-biased rates under the Negative condition and biased rates under the Non-negative condition, respectively, for Prompt 1 (left) and Prompt 2 (right). Color intensity indicates response rate and bubble size the number of valid cases; rows denote models and columns sensitive-attribute categories.}
    \label{fig:unknown_flow}
\end{figure*}

Figure \ref{fig:unknown_flow} indicates that the sensitive attribute categories were more pronounced than the models. In particular, high response rates were consistently observed for the categories "Race + SES," "Race + gender," "SES," and "Race/ethnicity." With composite attributes, the social associations involved in judgment become more complex; by removing abstention, a tendency to rely on the salient stereotypical associations within them may have come to the surface. This trend remained stable across the two types of prompts; the correlation coefficients between the results of Prompt 1 and Prompt 2 were 0.956 under the Negative condition and 0.889 under the Non-negative condition (Details in Table \ref{tab:pattern_stability} in Appendix). The median absolute difference was 2.10 and 1.80 percentage points, respectively, suggesting that the response patterns observed for each category did not depend heavily on specific prompt formulations. In the Qwen3 and Gemma3 families, as model size increased, there was a tendency for anti-biased response rates under the Negative condition and biased response rates under the Non-negative condition to rise across many categories. In Prompt 1 of SmolLM2-1.7B, the bubbles are conspicuously diminutive across multiple categories. This is due to a reduced number of "Unknown" responses in the three-choice format. In addition, the forced-choice response patterns are largely preserved after INT4 quantization, as Table \ref{tab:pattern_stability} in the Appendix \ref{sec:appendix_b}. 

\section{Interpretation of Findings}
The results show that the effects of model size and quantization extend beyond QA accuracy. Similar task performance did not always correspond to similar fairness-related behavior, with some configurations showing substantial changes in biased responses and abstention despite relatively small accuracy changes. This is consistent with prior work showing that compression can affect behavioral and safety dimensions differently from standard performance metrics \citep{xu_beyond_2024}.

The multi-prompt analysis further showed that these trends were not always robust to prompt formulation. While configurations such as SmolLM2 1.7B under INT8 Dynamic and Qwen3 1.7B under INT4 retained the same directional pattern across all five prompts, many configurations changed direction across formulations. This supports the use of multiple prompt versions when evaluating social bias \citep{hida_social_2025}.

Finally, the forced-choice experiment showed that the availability of an ``Unknown'' option can affect observed bias. When abstention was removed, previously ``Unknown'' responses were redistributed into biased or anti-biased choices. Together with the association between lower biased-response rates and increased over-abstention, this suggests that apparently lower bias can sometimes reflect greater abstention rather than a shift towards anti-biased judgments. Overall, fairness evaluation of quantized models should consider accuracy, biased responses, abstention, and prompt formulation jointly.

\section{Conclusion}
We investigated how model scale and quantization affect social bias and abstention in QA. Accuracy increased with model size, whereas bias showed no simple scaling trend and quantization effects varied across models. Lower biased-response rates were often accompanied by increased abstention. Under forced choice, category were more pronounced than model differences and remained largely stable across prompts and after INT4 quantization. These results highlight the importance of considering abstention, prompt formulation, and answer-option design in fairness evaluation.

\section*{Limitations}
This study has two main limitations. First, evaluation is limited to the BBQ, which may restrict generalizability to other bias benchmarks. Second, the analysis covers only relatively lightweight models in three families. Nevertheless, multiple model sizes and quantization settings were evaluated, providing some diversity within the study scope.

\section*{Ethical Considerations}
This study evaluates social bias using BBQ, which contains examples involving sensitive attributes and socially stereotyped associations. Such examples are used solely for the purpose of measuring model behavior and should not be interpreted as endorsing or validating the stereotypes represented in the benchmark. Our analysis is based on an existing public benchmark and does not involve human participants or
the collection of personal data.

\section*{Acknowledgments}
The authors would like to thank the EPSRC National Edge AI Hub at the University of Hull for their support throughout this research.


\bibliography{custom}

@inproceedings{xu_beyond_2024,
	address = {Miami, Florida, USA},
	title = {Beyond {Perplexity}: {Multi}-dimensional {Safety} {Evaluation} of {LLM} {Compression}},
	shorttitle = {Beyond {Perplexity}},
	url = {https://aclanthology.org/2024.findings-emnlp.901/},
	doi = {10.18653/v1/2024.findings-emnlp.901},
	urldate = {2026-09-25},
	booktitle = {Findings of the {Association} for {Computational} {Linguistics}: {EMNLP} 2024},
	publisher = {Association for Computational Linguistics},
	author = {Xu, Zhichao and Gupta, Ashim and Li, Tao and Bentham, Oliver and Srikumar, Vivek},
	editor = {Al-Onaizan, Yaser and Bansal, Mohit and Chen, Yun-Nung},
	month = nov,
	year = {2024},
	pages = {15359--15396},
}

@inproceedings{hida_social_2025,
	address = {Suzhou, China},
	title = {Social {Bias} {Evaluation} for {Large} {Language} {Models} {Requires} {Prompt} {Variations}},
	isbn = {979-8-89176-335-7},
	url = {https://aclanthology.org/2025.findings-emnlp.783/},
	doi = {10.18653/v1/2025.findings-emnlp.783},
	urldate = {2026-09-25},
	booktitle = {Findings of the {Association} for {Computational} {Linguistics}: {EMNLP} 2025},
	publisher = {Association for Computational Linguistics},
	author = {Hida, Rem and Kaneko, Masahiro and Okazaki, Naoaki},
	editor = {Christodoulopoulos, Christos and Chakraborty, Tanmoy and Rose, Carolyn and Peng, Violet},
	month = nov,
	year = {2025},
	pages = {14507--14530},
}

@inproceedings{rath_quantization_2026,
	title = {Quantization {Undoes} {Alignment}: {Bias} {Emergence} in {Compressed} {LLMs} {Across} {Models} and {Precision} {Levels}},
	shorttitle = {Quantization {Undoes} {Alignment}},
	url = {http://arxiv.org/abs/2605.15208},
	doi = {10.1109/CloudSummit68932.2026.00016},
	urldate = {2026-09-04},
	booktitle = {2026 {IEEE} {Cloud} {Summit}},
	author = {Rath, Plawan Kumar and Maliakkal, Rahul},
	month = jun,
	year = {2026},
	note = {arXiv:2605.15208 [cs.LG]},
	pages = {57--63},
}

@inproceedings{goncalves_understanding_2023,
	address = {Singapore},
	title = {Understanding the {Effect} of {Model} {Compression} on {Social} {Bias} in {Large} {Language} {Models}},
	url = {https://aclanthology.org/2023.emnlp-main.161/},
	doi = {10.18653/v1/2023.emnlp-main.161},
	urldate = {2026-09-04},
	booktitle = {Proceedings of the 2023 {Conference} on {Empirical} {Methods} in {Natural} {Language} {Processing}},
	publisher = {Association for Computational Linguistics},
	author = {Gonçalves, Gustavo and Strubell, Emma},
	editor = {Bouamor, Houda and Pino, Juan and Bali, Kalika},
	month = dec,
	year = {2023},
	pages = {2663--2675},
}

@inproceedings{tal_fewer_2022,
	address = {Seattle, Washington},
	title = {Fewer {Errors}, but {More} {Stereotypes}? {The} {Effect} of {Model} {Size} on {Gender} {Bias}},
	shorttitle = {Fewer {Errors}, but {More} {Stereotypes}?},
	url = {https://aclanthology.org/2022.gebnlp-1.13/},
	doi = {10.18653/v1/2022.gebnlp-1.13},
	urldate = {2026-09-04},
	booktitle = {Proceedings of the 4th {Workshop} on {Gender} {Bias} in {Natural} {Language} {Processing} ({GeBNLP})},
	publisher = {Association for Computational Linguistics},
	author = {Tal, Yarden and Magar, Inbal and Schwartz, Roy},
	editor = {Hardmeier, Christian and Basta, Christine and Costa-jussà, Marta R. and Stanovsky, Gabriel and Gonen, Hila},
	month = jul,
	year = {2022},
	pages = {112--120},
}

@misc{kraft_social_2026,
	title = {Social {Bias} in {Popular} {Question}-{Answering} {Benchmarks}},
	url = {http://arxiv.org/abs/2505.15553},
	doi = {10.48550/arXiv.2505.15553},
	urldate = {2026-09-04},
	publisher = {arXiv},
	author = {Kraft, Angelie and Simon, Judith and Schimmler, Sonja},
	month = jan,
	year = {2026},
	note = {arXiv:2505.15553 [cs.CL]},
}

@article{chen_fairquanti_2025,
	title = {{FairQuanti}: {Enhancing} {Fairness} in {Deep} {Neural} {Network} {Quantization} via {Neuron} {Role} {Contribution}},
	volume = {28},
	issn = {2471-2566},
	shorttitle = {{FairQuanti}},
	url = {https://dl.acm.org/doi/10.1145/3744560},
	doi = {10.1145/3744560},
	number = {3},
	urldate = {2026-09-04},
	journal = {ACM Transactions on Privacy and Security},
	author = {Chen, Jinyin and Cao, Zhiqi and Wang, Xiaojuan and Zheng, Haibin and Ming, Zhaoyan and Zheng, Yayu},
	month = aug,
	year = {2025},
	pages = {37:1--37:28},
}

@misc{ferrara_quantibias_2026,
	title = {{QuantiBias}: {Benchmarking} {Quantization}-{Induced} {Bias} in {LLMs}},
	shorttitle = {{QuantiBias}},
	url = {http://arxiv.org/abs/2607.21063},
	doi = {10.48550/arXiv.2607.21063},
	urldate = {2026-09-04},
	publisher = {arXiv},
	author = {Ferrara, Emilio},
	month = jul,
	year = {2026},
	note = {arXiv:2607.21063 [cs.CL]},
}

@article{proskurina_fair-gptq_2026,
	title = {Fair-{GPTQ}: {Bias}-{Aware} {Quantization} for {Large} {Language} {Models}},
	copyright = {Creative Commons Attribution 4.0 International},
	shorttitle = {Fair-{GPTQ}},
	url = {https://arxiv.org/abs/2509.15206},
	doi = {10.48550/ARXIV.2509.15206},
	urldate = {2026-09-04},
	publisher = {arXiv},
	author = {Proskurina, Irina and Metzler, Guillaume and Velcin, Julien},
	month = jul,
	year = {2026},
	note = {Version Number: 3},
}

@misc{hua_investigating_2026,
	title = {Investigating {Social} {Bias} {Changes} in {Quantized} {Language} {Models}},
	url = {https://arxiv.org/abs/2602.06181v2},
	language = {en},
	urldate = {2026-09-04},
	journal = {arXiv.org},
	author = {Hua, Stanley Z. and Lotfi, Sanae and Chen, Irene Y.},
	month = feb,
	year = {2026},
}

@article{marcuzzi_how_2025,
	title = {How {Quantization} {Shapes} {Bias} in {Large} {Language} {Models}},
	copyright = {Creative Commons Attribution 4.0 International},
	url = {https://arxiv.org/abs/2508.18088},
	doi = {10.48550/ARXIV.2508.18088},
	urldate = {2026-09-04},
	publisher = {arXiv},
	author = {Marcuzzi, Federico and Ning, Xuefei and Schwartz, Roy and Gurevych, Iryna},
	year = {2025},
	note = {Version Number: 3},
}

@misc{crawford_trouble_2017,
	title = {The {Trouble} with {Bias} - {NIPS} 2017 {Keynote} - {Kate} {Crawford} \#{NIPS2017}},
	url = {https://www.youtube.com/watch?v=fMym_BKWQzk},
	urldate = {2026-09-04},
	author = {Crawford},
	month = dec,
	year = {2017},
}

@inproceedings{baldini_your_2022,
	address = {Dublin, Ireland},
	title = {Your fairness may vary: {Pretrained} language model fairness in toxic text classification},
	shorttitle = {Your fairness may vary},
	url = {https://aclanthology.org/2022.findings-acl.176/},
	doi = {10.18653/v1/2022.findings-acl.176},
	urldate = {2026-09-03},
	booktitle = {Findings of the {Association} for {Computational} {Linguistics}: {ACL} 2022},
	publisher = {Association for Computational Linguistics},
	author = {Baldini, Ioana and Wei, Dennis and Natesan Ramamurthy, Karthikeyan and Singh, Moninder and Yurochkin, Mikhail},
	editor = {Muresan, Smaranda and Nakov, Preslav and Villavicencio, Aline},
	month = may,
	year = {2022},
	pages = {2245--2262},
}

@inproceedings{parrish_bbq_2022,
	address = {Dublin, Ireland},
	title = {{BBQ}: {A} hand-built bias benchmark for question answering},
	shorttitle = {{BBQ}},
	url = {https://aclanthology.org/2022.findings-acl.165/},
	doi = {10.18653/v1/2022.findings-acl.165},
	urldate = {2026-09-03},
	booktitle = {Findings of the {Association} for {Computational} {Linguistics}: {ACL} 2022},
	publisher = {Association for Computational Linguistics},
	author = {Parrish, Alicia and Chen, Angelica and Nangia, Nikita and Padmakumar, Vishakh and Phang, Jason and Thompson, Jana and Htut, Phu Mon and Bowman, Samuel R.},
	editor = {Muresan, Smaranda and Nakov, Preslav and Villavicencio, Aline},
	month = may,
	year = {2022},
	pages = {2086--2105},
}

@inproceedings{ma_mitigating_2023,
	title = {Mitigating {Bias} for {Question} {Answering} {Models} by {Tracking} {Bias} {Influence}},
	copyright = {Creative Commons Attribution 4.0 International},
	url = {https://arxiv.org/abs/2310.08795},
	doi = {10.48550/ARXIV.2310.08795},
	urldate = {2026-09-03},
	publisher = {arXiv},
	author = {Ma, Mingyu Derek and Kao, Jiun-Yu and Gupta, Arpit and Lin, Yu-Hsiang and Zhao, Wenbo and Chung, Tagyoung and Wang, Wei and Chang, Kai-Wei and Peng, Nanyun},
	year = {2023},
	note = {Version Number: 2},
}

@inproceedings{li_unqovering_2020,
	address = {Online},
	title = {{UNQOVERing} {Stereotyping} {Biases} via {Underspecified} {Questions}},
	url = {https://aclanthology.org/2020.findings-emnlp.311/},
	doi = {10.18653/v1/2020.findings-emnlp.311},
	urldate = {2026-09-03},
	booktitle = {Findings of the {Association} for {Computational} {Linguistics}: {EMNLP} 2020},
	publisher = {Association for Computational Linguistics},
	author = {Li, Tao and Khashabi, Daniel and Khot, Tushar and Sabharwal, Ashish and Srikumar, Vivek},
	editor = {Cohn, Trevor and He, Yulan and Liu, Yang},
	month = nov,
	year = {2020},
	pages = {3475--3489},
}

@inproceedings{manduru2026beyond,
  title={Beyond Bias Scores: Unmasking Vacuous Neutrality in Small Language Models},
  author={Manduru, Sumanth and Domeniconi, Carlotta},
  booktitle={Proceedings of the 19th Conference of the European Chapter of the Association for Computational Linguistics (Volume 4: Student Research Workshop)},
  pages={685--714},
  year={2026}
}

@inproceedings{parrish2022bbq,
  title={BBQ: A hand-built bias benchmark for question answering},
  author={Parrish, Alicia and Chen, Angelica and Nangia, Nikita and Padmakumar, Vishakh and Phang, Jason and Thompson, Jana and Htut, Phu Mon and Bowman, Samuel R},
  booktitle={Findings of the Association for Computational Linguistics: ACL 2022},
  pages={2086--2105},
  year={2022}
}

@misc{qwen3technicalreport,
      title={Qwen3 Technical Report}, 
      author={Qwen Team},
      year={2025},
      eprint={2505.09388},
      archivePrefix={arXiv},
      primaryClass={cs.CL},
      url={https://arxiv.org/abs/2505.09388}, 
}

@article{hida2024social,
  title={Social bias evaluation for large language models requires prompt variations},
  author={Hida, Rem and Kaneko, Masahiro and Okazaki, Naoaki},
  journal={arXiv preprint arXiv:2407.03129},
  year={2024}
}

@article{or2025torchao,
  title={Torchao: Pytorch-native training-to-serving model optimization},
  author={Or, Andrew and Jain, Apurva and Vega-Myhre, Daniel and Cai, Jesse and Hernandez, Charles David and Zheng, Zhenrui and Guessous, Driss and Kuznetsov, Vasiliy and Puhrsch, Christian and Saroufim, Mark and others},
  journal={arXiv preprint arXiv:2507.16099},
  year={2025}
}

@article{allal2025smollm2,
  title={SmolLM2: When Smol Goes Big--Data-Centric Training of a Small Language Model},
  author={Allal, Loubna Ben and Lozhkov, Anton and Bakouch, Elie and Bl{\'a}zquez, Gabriel Mart{\'\i}n and Penedo, Guilherme and Tunstall, Lewis and Marafioti, Andr{\'e}s and Kydl{\'\i}{\v{c}}ek, Hynek and Lajar{\'\i}n, Agust{\'\i}n Piqueres and Srivastav, Vaibhav and others},
  journal={arXiv preprint arXiv:2502.02737},
  year={2025}
}

@article{sood2026punching,
  title={Punching Above Their Weight: Classification-Head Fine-Tuning of Tiny Language Models (TLMs) for Verifiable Multiple-Choice Tasks},
  author={Sood, Bhavesh and Savelka, Jaromir},
  journal={arXiv preprint arXiv:2607.03801},
  year={2026}
}

@article{geetha2026tiny,
  title={Tiny Language Model-Based Embedded Decision Engine for Resource-Constrained Intelligent Wireless Sensor Networks},
  author={Geetha, K},
  journal={Journal of Wireless Intelligence and Spectrum Engineering},
  volume={3},
  number={3},
  pages={11--23},
  year={2026}
}

@article{liu2024mobilellm,
  title={Mobilellm: Optimizing sub-billion parameter language models for on-device use cases},
  author={Liu, Zechun and Zhao, Changsheng and Iandola, Forrest and Lai, Chen and Tian, Yuandong and Fedorov, Igor and Xiong, Yunyang and Chang, Ernie and Shi, Yangyang and Krishnamoorthi, Raghuraman and others},
  journal={arXiv preprint arXiv:2402.14905},
  year={2024}
}

@article{dettmerslambda2022llm,
  title={LLM. int8 (): 8-bit Matrix Multiplication for Transformers at Scale},
  author={Dettmers$\lambda$, Tim and Lewis, Mike and Belkada, Younes and Zettlemoyer, Luke and Face, Hugging and Paris-Saclay, ENS},
  journal={arXiv preprint arXiv:2208.07339},
  year={2022}
}

@article{team2025gemma,
  title={Gemma 3 technical report},
  author={Team, Gemma and Kamath, Aishwarya and Ferret, Johan and Pathak, Shreya and Vieillard, Nino and Merhej, Ramona and Perrin, Sarah and Matejovicova, Tatiana and Ram{\'e}, Alexandre and Rivi{\`e}re, Morgane and others},
  journal={arXiv preprint arXiv:2503.19786},
  year={2025}
}

@article{yao2022zeroquant,
  title={Zeroquant: Efficient and affordable post-training quantization for large-scale transformers},
  author={Yao, Zhewei and Yazdani Aminabadi, Reza and Zhang, Minjia and Wu, Xiaoxia and Li, Conglong and He, Yuxiong},
  journal={Advances in neural information processing systems},
  volume={35},
  pages={27168--27183},
  year={2022}
}

@article{haider2024phi,
  title={Phi-3 safety post-training: Aligning language models with a" break-fix" cycle},
  author={Haider, Emman and Perez-Becker, Daniel and Portet, Thomas and Madan, Piyush and Garg, Amit and Ashfaq, Atabak and Majercak, David and Wen, Wen and Kim, Dongwoo and Yang, Ziyi and others},
  journal={arXiv preprint arXiv:2407.13833},
  year={2024}
}

@article{lin2024awq,
  title={Awq: Activation-aware weight quantization for on-device llm compression and acceleration},
  author={Lin, Ji and Tang, Jiaming and Tang, Haotian and Yang, Shang and Chen, Wei-Ming and Wang, Wei-Chen and Xiao, Guangxuan and Dang, Xingyu and Gan, Chuang and Han, Song},
  journal={Proceedings of machine learning and systems},
  volume={6},
  pages={87--100},
  year={2024}
}

\appendix
\section{Additional Explanation of Methodology}
\label{sec:appendix_a}
\subsection{Computing Resources}
This study was conducted under the experimental conditions shown in the Table \ref{tab:execution_environment} below.
\begin{table}[H]
\centering
\small
\begin{tabular}{l|l}
\hline
Items         & Specs                    \\ \hline
CPU           & Intel Core i9-14900F     \\
GPU           & NVIDIA GeForce RTX 4090  \\
GPU Memory    & 24 GB                    \\
System Memory & 64 GB                    \\
OS            & Ubuntu 22.04.4 LTS (WSL2) \\
Python        & 3.11.15                  \\
PyTorch       & 2.11.0+cu126             \\
Accelerate    & 1.14.0                   \\
TorchAO       & 0.17.0+cu126            \\ \hline
\end{tabular}
\caption{Execution Environment}
\label{tab:execution_environment}
\end{table}

\subsection{Selected Models}
We selected three model families—Qwen3, Gemma-3, and SmolLM2—for evaluation. The reasons for this selection of Qwen3, Gemma-3, and SmolLM2 are as follows. (1) These model families are closely related to prior research on lightweight language models, making them well-suited for comparison with existing studies and for contextualizing our results. (2) Pre-trained checkpoints are available for free, allowing us to conduct experiments without relying on paid APIs or token-based services. (3) Since each model family includes models with different numbers of parameters, we can systematically compare the effects of model scale within the same family. (4) These models can apply different quantization settings under common experimental conditions and evaluate their effects. 

\section{Additional Results of Forced-Questioning Analysis}
\label{sec:appendix_b}
\begin{table}[H]
\centering
\resizebox{\columnwidth}{!}{%
\begin{tabular}{lll}
\hline
Comparison             & Negative     & Non-negative \\ \hline
Prompt 1 vs. Prompt 2  & 0.956 / 2.10 & 0.889 / 1.80 \\
FP vs. INT4 Prompt 1 & 0.975 / 1.16 & 0.983 / 0.95 \\
FP vs. INT4 Prompt 2 & 0.992 / 0.98 & 0.990 / 1.11 \\
\hline
\end{tabular}
}
\caption{Stability of category-level response patterns across prompt formulations and INT4 quantization. Values are Pearson correlation / median absolute difference.}
\label{tab:pattern_stability}
\end{table}

\end{document}